\documentclass[conference]{ieeeconf}
\IEEEoverridecommandlockouts

\usepackage{cite}
\usepackage{algorithmic}
\usepackage{textcomp}
\usepackage{pgfplots}
\usepackage{pgfplotstable}
\def\BibTeX{{\rm B\kern-.05em{\sc i\kern-.025em b}\kern-.08em
    T\kern-.1667em\lower.7ex\hbox{E}\kern-.125emX}}

\usepackage{xcolor}
\usepackage[T1]{fontenc} % Auswahl einer Schriftkodierung für europäische Sprachen
\usepackage[utf8]{inputenc} % Auswahl einer Textkodierung für europäische Sprachen
\usepackage[english]{babel} % Auswahl der Sprache ngerman oder english

\usepackage{amsmath} % assumes amsmath package installed
\usepackage{amsfonts}
\usepackage{amssymb}  % assumes amsmath package installed

\usepackage{booktabs} % Erstellen schöner Tabellen

\usepackage{dsfont}
\usepackage{graphicx}
\usepackage{subcaption} 

\usepackage{tikz}
\usepackage{pgfplots}
\pgfplotsset{compat=newest}
\usepgfplotslibrary{fillbetween}

\usepackage{epsfig} % for postscript graphics files
\usepackage{times} % assumes new font selection scheme installed
\usepackage{siunitx}
\usepackage{calc}
\usepackage[hyphens]{url}
\usepackage{svg}
\renewcommand{\baselinestretch}{0.924}
\usepackage{adjustbox}

\usepackage{tabularx}
\usepackage{bbm}
\usepackage{dsfont}
\title{\LARGE \textbf{Transfer Learning Study of Motion Transformer-based\\Trajectory Predictions*
}

\author{Lars Ullrich, Alex McMaster and Knut Graichen% <-this % stops a space
	\thanks{*This research is accomplished within the project ”AUTOtechagil” (FKZ 01IS22088Y). We acknowledge the financial support for the project by the Federal Ministry of Education and Research of Germany (BMBF).}
	\thanks{The authors are with the Chair of Automatic Control, Friedrich-Alexander-Universität Erlangen-Nürnberg (FAU), Germany {\tt\footnotesize \{lars.ullrich, alex.r.mcmaster, knut.graichen\}@fau.de}}%
}

}

\usepackage{hologo}
\usepackage{listings}
\begin{document}

% for arXiv publication with appropriate copyright notice
\twocolumn[
\begin{@twocolumnfalse}
\Huge {IEEE copyright notice} \\ \\
\large {\copyright\ 2024 IEEE. Personal use of this material is permitted. Permission from IEEE must be obtained for all other uses, in any current or future media, including reprinting/republishing this material for advertising or promotional purposes, creating new collective works, for resale or redistribution to servers or lists, or reuse of any copyrighted component of this work in other works.} \\ \\

{\Large Published in \emph{2024 35th IEEE Intelligent Vehicles Symposium (IV)}, Jeju Island, Korea, June 2 - 5, 2024.} \\ \\

Cite as:

\vspace{0.1cm}
\noindent\fbox{%
	\parbox{\textwidth}{%
		L.~Ullrich, A.~McMaster, and K.~Graichen, ``Transfer Learning Study of Motion Transformer-based Trajectory Predictions,''
		in \emph{2024 35th IEEE Intelligent Vehicles Symposium (IV)}, Jeju Island, Korea, 2024, pp. 110--117, doi: 10.1109/IV55156.2024.10588422.
	}%
}
\vspace{2cm}

\end{@twocolumnfalse}
]

\noindent\begin{minipage}{\textwidth}
	
\hologo{BibTeX}:
\footnotesize
\begin{lstlisting}[frame=single]
@inproceedings{ullrich2024transfer,
	author={Ullrich, Lars and McMaster, Alex and Graichen, Knut},
	booktitle={2024 35th IEEE Intelligent Vehicles Symposium (IV)},
	title={Transfer Learning Study of Motion Transformer-based Trajectory Predictions},
	address={Jeju Island, Korea},
	year={2024},
	pages={110--117},
	doi={10.1109/IV55156.2024.10588422},
	publisher={IEEE}
}
\end{lstlisting}
\end{minipage}
    
\thispagestyle{empty}
\pagestyle{empty}
\bstctlcite{IEEEexample:BSTcontrol}

\maketitle

\begin{abstract}
Trajectory planning in autonomous driving is highly dependent on predicting the emergent behavior of other road users. Learning-based methods are currently showing impressive results in simulation-based challenges, with transformer-based architectures technologically leading the way. Ultimately, however, predictions are needed in the real world. In addition to the shifts from simulation to the real world, many vehicle- and country-specific shifts, i.e. differences in sensor systems, fusion and perception algorithms as well as traffic rules and laws, are on the agenda. Since models that can cover all system setups and design domains at once are not yet foreseeable, model adaptation plays a central role. Therefore, a simulation-based study on transfer learning techniques is conducted on basis of a transformer-based model. Furthermore, the study aims to provide insights into possible trade-offs between computational time and performance to support effective transfers into the real world.
\end{abstract}

\section{INTRODUCTION}

The anticipation of trajectories of fellow road users constitutes a fundamental prerequisite of informed trajectory planning, thereby affording a basis for dependable local decision-making of autonomous vehicles in the real world, where space is collaboratively shared. This task of predicting other road users trajectory has received significant attention in recent years \cite{shi2022motion, lin2023eda, gan2023mgtr}. Similar to advancements in the field of perception \cite{geiger2013vision, sun2020scalability, caesar2020nuscenes}, trajectory prediction witnessed the emergence of numerous datasets and challenges \cite{chang2019argoverse, houston2021one, ettinger2021large}, that now serve as a solid research foundation. These challenges have evolved to address increasingly complex and realistic problems, such as joint predictions of multiple road users \cite{ettinger2021large}. Moreover, there is also a current expansion in the examination of open-loop predictions to encompass closed-loop predictions \cite{caesar2021nuplan}, reflecting a more real world application scenario.

Thereby, traditional physical approaches are not suitable for targeted prediction horizons of up to eight seconds\textsuperscript{1} due to context disregard and missing behavioral knowledge. In addition, rule-based procedures are not able to cope with the multitude of corner cases that exist in the real world \cite{chen2023end}. In contrast, learning-based approaches are able to incorporate contextual information and learn to anticipate various behaviors \cite{shi2022motion}, allowing them to perform well even with prediction horizons up to eight seconds. Learning-based approaches are also promising with regard to handling open long-tail distributions of situations, for example, through generalization \cite{chen2023end}. 

Besides the major challenge of explainability in AI, another significant concern is the impact of changes in the underlying data generating process, known as distribution shifts, as they have a strong impact on the performance \cite{filos2020can, chen2023end}. Although current efforts in trajectory prediction research effectively tackle the inherent complexity of real world tasks within a controlled research setting, the aspect of transferability is often overlooked. Nevertheless, effective transferability is crucial for scalability and, consequently, for real world relevance.

Significant differences are imaginable across operational design domains (ODD), encompassing factors like left and right-hand traffic, country-specific traffic signs, and diverse traffic regulations. Additionally, vehicles vary widely in terms of individual equipment, including sensor technology and subsequent data fusion. Generating datasets for every conceivable situation, ODD, and vehicle configuration would result in vast amounts of data and substantial costs. Mitigating this challenge through domain and vehicle generalization is also questionable in accordance to \cite{codevilla2019exploring}. Rather a more pragmatic objective is transferring the knowledge across ODDs and vehicle configurations using advanced transfer learning techniques. This approach acknowledges the impracticality of exhaustive dataset generation while striving to reserve generalization capabilities for corner case scenarios within specific ODDs and vehicle configurations.
% considering e.g. generalization limitations of imitation learning \cite{codevilla2019exploring}
In this paper, a transfer learning study is carried out in order to shed more light onto relevant aspects related to transfer learning. For safety reasons, the study is conducted in the simulation environment CarMaker\footnote{\url{https://ipg-automotive.com/en/products-solutions/software/carmaker/}} which is widely used in the automotive industry. The constructed CarMaker Dataset (CMD) serves as the target and the Waymo Open Motion Dataset (WOMD) \cite{ettinger2021large} as the source of the transfer learning study. While plenty of learning-based methods exist in the field of trajectory prediction, transformer-based approaches \cite{shi2022motion, shi2024mtrpp, lin2023eda, gan2023mgtr}, are leading the way, as can be seen e.g. from the Waymo Leader Board\footnote{\url{https://waymo.com/open/challenges/2023/motion-prediction/}}. However, transformer approaches used in motion prediction differ significantly from transformers \cite{vaswani2017attention} originally used in machine translation due to the different nature of task.

Unlike machine translation, which operates in a specific alphabet space, trajectory prediction concerns a temporal-spatial task. Moreover, whereas machine translation only requires one translation, different possibilities akin modalities and their probabilities as well as uncertainties within modalities are of significant importance for safe and reliable trajectory planning. Therefore, direct conclusions about transformers for motion predictions cannot be drawn from transfer learning studies of classical transformers \cite{raffel2020exploring}. 
Among the many existing models, the baseline motion transformer (MTR) \cite{shi2022motion} is selected as basis of the study. Our contribution is the investigation of transfer learning capabilities based on the MTR. Thereby, insights into the trade-offs between computation time and accuracy are provided. The knowledge gained aims to facilitate a timely transition of remarkable performances from academic challenges to the real world.

The paper is structured as follows: the motion transformer, which serves as the basis for the study, is outlined in Section \ref{motion_transformer}. The methodology of the transfer learning study is elaborated in Section \ref{transfer_learning}. The evaluation is described in Section \ref{analysis} along with the corresponding results. The results and, in particular, possible future research directions are discussed in Section \ref{discussion}. Finally, findings are summarized in Section \ref{conclusion}.
\section{MOTION TRANSFORMER }\label{motion_transformer}

The Motion Transformer (MTR) \cite{shi2022motion} is a model architecture that has achieved state of the art results in trajectory prediction for autonomous driving, scoring 1st on the Waymo Motion Prediction Challenge 2022. The framework of MTR combines advantages of previous leading paradigms of goal-based and direct-regression methods by jointly optimizing global intention and local movement refinement. Therefore, a transformer encoder-decoder architecture is introduced with a novel motion decoder that leverages on learnable motion queries for efficient and accurate predictions.

\subsection{Related Work and Architectural Overview}\label{motion_transformer_A}
 % \textbf{Idea: turn order, first comarision, second high level description, third ref to below. Finally ref for more information to paper.}

In the field of multimodal trajectory predictions, a set $\Gamma = \{\Gamma_k\}_{k=1}^{K}$ of $K$ predictions $\Gamma_k = \{\hat{\boldsymbol{\tau}}_k, c_k\}$ is considered. Thereby, $\hat{\boldsymbol{\tau}}_k = \{\hat{\boldsymbol{\tau}}_{k,t}\}_{t=1}^{T}$ represents the trajectory of modality $k$ for time steps from 1 to $T$, and $c_k$ denotes the corresponding scalar confidence value associated with the trajectory. For evaluation, various performance metrics are used in this context. Among others, the established metrics minimum Average Displacement Error (minADE), minimum Final Displacement Error (minFDE) and the miss Rate are used \cite{ettinger2021large}. The minADE of an agent $a$ is calculated as
\begin{equation}\label{eq:minADE}
    \mathrm{minADE}(a) = \min_{k} \frac{1}{T} \sum_{t}^{T} \| \boldsymbol{\tau}_{t}^{a} - \hat{\boldsymbol{\tau}}_{k,t}^{a} \|_2,
\end{equation}
the L2 norm between the ground truth trajectory $\boldsymbol{\tau}$ and the closest predicted trajectory $\hat{\boldsymbol{\tau}}$. In comparison, the minFDE specifically focuses on the final time step $T$ of the prediction horizon and is computed as follows
\begin{equation}\label{eq:minFDE}
    \mathrm{minFDE}(a) = \min_{k} \| \boldsymbol{\tau}_{T}^{a} - \hat{\boldsymbol{\tau}}_{k,T}^{a} \|_2.
\end{equation}
Furthermore, the miss rate is defined by the ratio of the total number of misses to the total number of multimodal predictions. A multimodal prediction set $\Gamma$ is classified as a miss if none of the $K$ trajectory predictions $\boldsymbol{\hat{\tau}}_{k}$ match the ground truth trajectory $\boldsymbol{\tau}$. The evaluation of the predicted trajectories follows the methodology outlined in \cite{ettinger2021large}, using the binary indicator function
\begin{align}\label{eq:isMatch}
	\begin{split}
        \mathrm{isMatch}(\boldsymbol{\tau}_t, \hat{\boldsymbol{\tau}}_t) &=  \mathds{1} [ x_{k,t} < \xi_{\mathrm{long}} ]\cdot \mathds{1} [ y_{k,t} < \xi_{\mathrm{lat}}], \\
        [x_{k,t}, y_{k,t}] &:= (\boldsymbol{\tau}_t - \hat{\boldsymbol{\tau}}_{k,t}) \cdot \boldsymbol{R}_t,
	\end{split}
\end{align}
where $\xi_{\mathrm{long}}$ and $\xi_{\mathrm{lat}}$ are longitudinal and lateral thresholds w.r.t. the ground truth heading of the agent $a$ at time $t$. In this context, trajectories are transformed into the respective coordinates via the corresponding 2D rotation matrix $\boldsymbol{R}_t$. The threshold values increase over the prediction horizon and are scaled depending on the agents present velocity to ensure a fair evaluation across diverse scenarios \cite{ettinger2021large}.

Beyond these metrics, the Waymo Motion Prediction Challenge relies strongly on the mean Average Precision (mAP), that is inspired and aligned with the definition in the field of object detection, exemplified by the Pascal Visual Object Classes (VOC) Challenge \cite{everingham2010pascal}. In general, there is a fixed number of semantic categories representing different generic trajectory shapes for which the Average Precision (AP) per category is first calculated and then averaged over all categories to obtain the mAP. The AP metric uses (\ref{eq:isMatch}) and the confidence value $c_k$ for a confusion matrix-like classification of the predictions in order to subsequently determine precision and recall. Again, AP aligns with object detection setups allowing only one true positive, namely the prediction with the highest confidence value, that is a match. Finally, the AP is calculated as the area under the precision-recall curve and averaged over the categories to obtain the mAP \cite{ettinger2021large}.

The metrics presented are used to compare the different models, as displayed in the Table \ref{tab:performance_metrics}. Notably, the methods MTR++ \cite{shi2024mtrpp}, IAIR+ \cite{kang2023iair}, and GTR-R36 \cite{liu2023transformer}, which took first to third place in the Waymo Challenge 2023\footnote{\url{https://waymo.com/open/challenges/2023/motion-prediction/}}, are all derived from the MTR framework \cite{shi2022motion}. Additionally, other high-performing approaches of the latest reasearch such as EDA \cite{lin2023eda} and MGTR \cite{gan2023mgtr} complement the MTR framework as well. This illustrates that the MTR framework has quickly established itself as a baseline and is fundamental to the latest research. Hence, we have chosen MTR as baseline for our studies, as new models typically make only minor adjustments to the framework.

\begin{table}[]
  \centering
  \caption{Performance metrics comparison.}
  \label{tab:performance_metrics}
  \begin{tabular}{p{1.66cm}p{1.05cm}p{0.88cm}p{0.88cm}p{0.88cm}p{0.88cm}}
    \toprule
    Method & Note & mAP$\uparrow$ & mADE$\downarrow$ & minFDE$\downarrow$ & missRate$\downarrow$ \\
    \midrule
    MTR \cite{shi2022motion} & 1st 2022 & 0.4129 & 0.6050 & 1.2207 & 0.1351 \\
    GTR-R36 \cite{liu2023transformer} & 3rd 2023 & 0.4255 & 0.6005 & 1.2225 & 0.1330 \\
    IAIR+ \cite{kang2023iair} & 2nd 2023 & 0.4347 & 0.5783 & \textbf{1.1679} & 0.1238 \\
    MTR++ \cite{shi2024mtrpp} & 1st 2023 & 0.4329 & 0.5906 & 1.1939 & 0.1298 \\
    EDA \cite{lin2023eda} & newer & 0.4401 & \textbf{0.5718} & 1.1702 & \textbf{0.1169} \\
    MGTR \cite{gan2023mgtr} & newer & \textbf{0.4505} & 0.5918 & 1.2135 & 0.1298 \\
    \bottomrule
  \end{tabular}
\end{table}

\begin{figure}[]
	\centering	
	\includegraphics[scale=0.22]{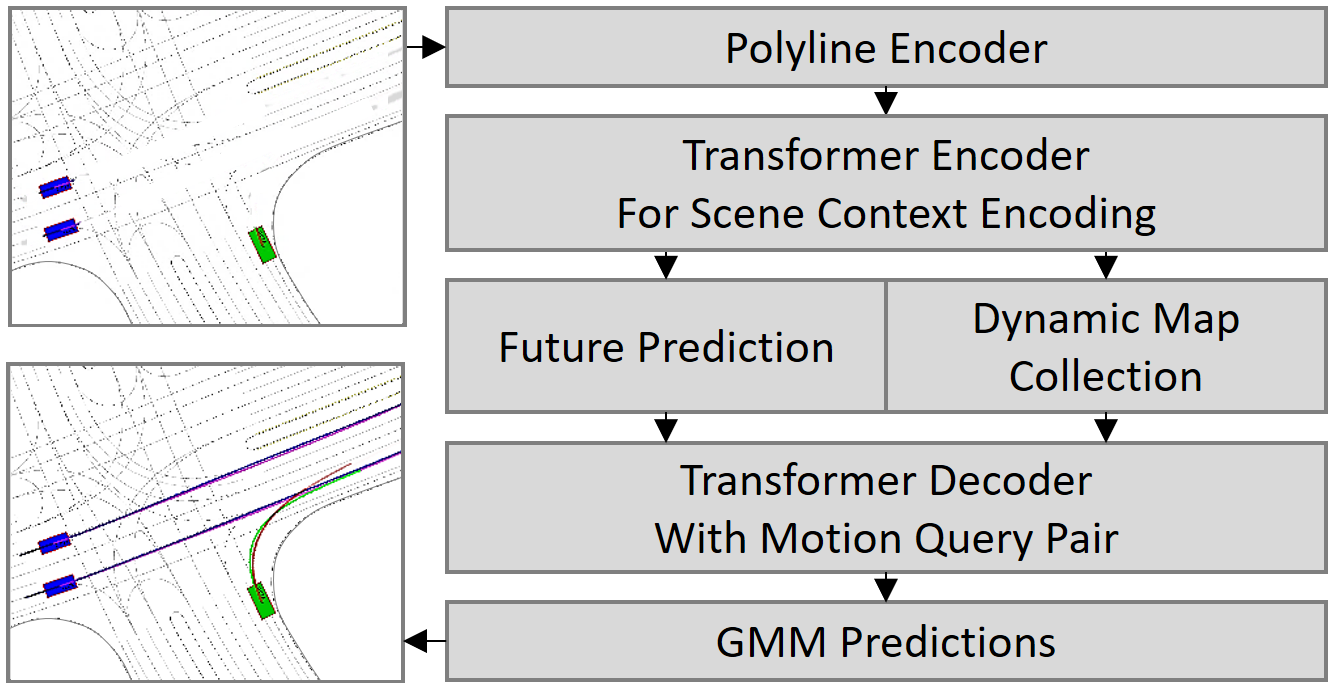}
	\caption{Simplified representation of the MTR architecture.}
	\label{fig:MTR}
\end{figure}

The architecture of the MTR is depicted in simplified form in Figure \ref{fig:MTR}. In the following subsections, the information encoding and decoding is described in more detail. 

% s[width=\linewidth]

\subsection{Information Encoding}
The MTR framework \cite{shi2022motion} builds upon a previously used vectorized \cite{gao2020vectornet} and agent-centric normalized \cite{zhao2021tnt} representation, portraying the trajectory history of road users, referred to as agents, and road map information as polylines $S_i$ with $i\in\{a,m\}$ for agents and map, respectively. The polylines are denoted as $S_i \in \mathbb{R}^{N_i \times P_i \times C_i}$, where $N_i$ represents the number of agents and the number of map polylines, $P_i$ signifies the number of historical time steps in the trajectory or the number of polyline points, and $C_i$ indicates the number of state information (e.g. location, heading) for agents or the attributes of road features (e.g. location, road type) for the map. A PointNet-like \cite{qi2017pointnet} polyline encoder 
\begin{equation}\label{eq:PolyEncoder}
    F_i =  \phi ( \mathrm{MLP} \bigl( S_i \bigr)), \quad i\in\{a,m\},
\end{equation}
transforms polylines $S_i$ via a multilayer perceptron network $\mathrm{MLP}(\cdot)$, followed by max-pooling $\phi$ into agent and map features denoted as $ F_{i} \in \mathbb{R}^{N_i \times D}$, $i\in\{a,m\}$ and feature dimension $D$.

The concatenation of features $F_i$ yields input tokens $F_{c} \in \mathbb{R}^{(N_a + N_m) \times D}$ for the transformer encoder, which is responsible for scene context encoding. By using local self-attention, an efficient mechanism for local context encoding via dedicated token aggregation and encoding is implemented. The attention within each individual encoder layer is formally expressed as
\begin{align}\label{eq:EncoderMHSA}
	\begin{split}
        Q &= F^{j-1}_{c} + \mathrm{PE}(F^{j-1}_{c}), \\
        K &= \kappa(F^{j-1}_{c}) + \kappa(\mathrm{PE}(F^{j-1}_{c})),  \quad V =\kappa(F^{j-1}_{c})), \\
        F^{j}_{c} &= \mathrm{MHSA}(Q, K, V),
	\end{split}
\end{align}
here $\mathrm{MHSA}(Q,K,V)$ denotes the multi-head self-attention layer \cite{vaswani2017attention}, with $Q$, $K$, and $V$ representing the query, key, and value components, respectively. The sinusoidal positional encoding \cite{vaswani2017attention} is symbolized by $\mathrm{PE}(\cdot)$, while $\kappa(\cdot)$ specifically denotes the \textit{k}-nearest neighbor algorithm for closest token identification. Multiple layers constitute the transformer encoder, which generates token features $F'_{c} \in \mathbb{R}^{(N_a + N_m) \times D}$. Before being used in the decoder, these features are enhanced through context-based trajectory predictions for future time steps, e.i.
\begin{equation}\label{eq:EncoderPred}
    \mathrm{CTP}_{1:T} = \mathrm{MLP}( F'_a) \in \mathbb{R}^{N_a \times T \times 4},
\end{equation}
where $1:T$ denotes the time steps from 1 to $T$ of the trajectories, which consist of 2D positions and velocities. After featurizing generated future trajectories using the polyline encoder (\ref{eq:PolyEncoder}) and concatenating them with $F'_a$, an additional MLP processing is applied to update the agent tokens $F'_a$. Consequently, $F'_a$ now encompasses both historical and context-based future trajectories, that can be further processed by the decoder.

\subsection{Information Decoding}
The MTR framework introduces a novel motion query pair consisting of a static intention query and a dynamic searching query. The static intention query represents an efficient and effective improvement of goal-based methods \cite{gu2021densetnt} by generating $\mathcal{K}$ representative 2D intention goals $I \in \mathbb{R}^{\mathcal{K} \times 2}$ via K-means clustering across ground truth trajectories. Building upon intention goals, each representing an implicit motion mode, intention queries
\begin{equation}\label{eq:IntentionQuery}
Q_{I} = \mathrm{MLP}(\mathrm{PE}(I)) \in \mathbb{R}^{\mathcal{K}\times D}
\end{equation}
are generated by a learnable position embedding. Analogously, but dynamically adapted across decoder layers, the dynamic search query of the ($j+1$)-th decoder layer is formulated as
\begin{equation}\label{eq:DynamicSearchQuery}
Q^{j+1}_{S} = \mathrm{MLP}(\mathrm{PE}(Y^{j}_T)) \in \mathbb{R}^{\mathcal{K}\times D}.
\end{equation}
In this context, the dynamic search query is initialized by intention goals $I$ and progressively refined across each decoder layer using a set of endpoints $Y^{j}_T$ derived from predicted trajectories $Y^{j}_{1:T} = \{Y^{j}_i \in \mathbb{R}^{\mathcal{K}\times 2} | i = 1, \dots , T\}$ of the $j$-th decoder layer.
 
The motion query pair undergoes joint optimization and refinement across all decoder layers, while also considering scene context obtained through previous encoding. The composition of individual decoder layers is outlined in the following. 
First, the static intention query is employed as as position embeddings in a multi-head self-attention 
\begin{align}\label{eq:DecoderMHSA}
	\begin{split}
        Q &= K = C^{j-1} + Q_I, \quad V = C^{j-1}, \\
        C^{j}_{qc} &= \mathrm{MHSA}(Q, K, V),
	\end{split}
\end{align}
% \begin{align}\label{eq:DecoderMHSA}
% 	\begin{split}
%         C^{j}_{qc} = \mathrm{MHSA}\Bigl( &Q = C^{j-1} + Q_I, \\
%         &K = C^{j-1} + Q_I, \\
%         &V = C^{j-1} \Bigr),
% 	\end{split}
% \end{align}
where $C^{j-1} \in \mathbb{R}^{\mathcal{K} \times D}$ denotes query content features of the $(j-1)$-th decoder layer, $C^j_{qc}$ signifies updated query content, and $C^0$ is initialized with zeros. Thereafter, the dynamic search query is used as position embedding in the query $Q$ and key $K$ of two multi-head cross-attention modules (MHCA). One is dedicated to agent feature aggregation, formalized as
\begin{align}\label{eq:DecoderMHCAAAgent}
	\begin{split}
        Q &= [C^{j}_{qc}, Q^j_S], \\
        K &= [F'_a, \mathrm{PE}(F'_a)],  \quad V = F'_a,  \\
        C^{j}_{a} &= \mathrm{MHCA}(Q, K, V),
	\end{split}
\end{align}
with $[\cdot, \cdot]$ denoting concatenation. Thus, taking into account updated query content $C^{j}_{qc}$ as well as agent tokens $F'_a$ from encoding. The other MHCA, similarly, aggregates map features as
\begin{align}\label{eq:DecderMHCAMap}
	\begin{split}
        Q &= [C^{j}_{qc}, Q^j_S], \\
        K &= [\alpha(F'_a), \mathrm{PE}(\alpha(F'_a))], \quad V = F'_a,  \\
        C^{j}_{m} &= \mathrm{MHCA}(Q, K, V),
	\end{split}
\end{align}
where $\alpha(\cdot)$ signifies a dynamic map collection responsible for querying local region map features, thereby accounting for road map based behavior guidance. Subsequently, the MHCA outputs $C^{j}_{a}$ and $C^{j}_{m}$ are concatenated and MLP post-processed, yielding the query content $C^j \in \mathbb{R}^{\mathcal{K} \times D}$ of $j$-th decoder layer. 

In addition to the use of $C^j$ in the subsequent decoder layers, on this basis the multimodal predictions are generated, represented according to \cite{chai2019multipath}. Here, the $\mathcal{K}$ modalities are characterized in each future step $h$ within the prediction horizon $T$ by two-dimensional Gaussian components denoted by $\mathcal{N}_{1:\mathcal{K}}(\mu_x, \sigma_x; \mu_y, \sigma_y; \rho)$, each with individual means $\mu_x, \mu_y$, standard deviations $\sigma_x, \sigma_y,$ and correlation coefficient $\rho$. Together with the occurrence probabilities $p_{1:\mathcal{K}}$ of each intention goal $I$ corresponding to the $\mathcal{K}$ modalities, the predicted occurrence probability density of the agents spatial position at time step $h$ is given by
\begin{equation}\label{eq:Trajectory}
    P_{h}(o) = \sum_{k=1}^{\mathcal{K}} p_k \cdot  \mathcal{N}_{k}(o_k - \mu_x, \sigma_x; o_y - \mu_y, \sigma_y; \rho),
\end{equation}
with $o\in\mathbb{R}^2$ denoting the spatial position of the agent. The required parameters of the Gaussian components $\mathcal{N}_{1:\mathcal{K}}(\mu_x, \sigma_x; \mu_y, \sigma_y; \rho)$ as well as the intentions occurrence probability $p_{1:\mathcal{K}}$ is parameterized by $Z^{j}_{1:T} = \mathrm{MLP}(C^j) \in \mathbb{R}^{\mathcal{K}\times6}$. Using the centers of the Gaussian components, the predicted trajectories $Y^{j}_{1:T}$ of the $j$-th decoder layer can be extracted. These are in turn dynamically retrieved by the search query, such that a local movement refinement takes place over the decoder layers before trajectory predictions are ultimately returned by the MTR framework.

\subsection{Consolidation \& Motivation}
It is evident that the MTR framework \cite{shi2022motion} has quickly emerged as a foundation of current research in the field of vehicle motion prediction, as indicated in Table \ref{tab:performance_metrics}. In particular, the joint optimization of goal intention and local motion refinement based on a transformer architecture seems to offer a targeted integration of goal-based (e.g., \cite{gu2021densetnt}) and direct regression-based (e.g., \cite{ngiam2022scene}) methods. While this approach has been underpinned by further research as part of the Waymo Motion Prediction Challenge \cite{gan2023mgtr, lin2023eda, liu2023transformer, kang2023iair}, the transferability of such systems remains an open research question.

However, the importance of transferability and the inherent challenges in applying machine learning methods to sophisticated real-world tasks are well known and are addressed, for example, in dedicated academic challenges such as Shifts \cite{malinin2021shifts, malinin2022shifts}. In addition, related research areas have evolved in the field of artifical intelligence. These range from explainable and trustworthy AI to more tangible and problem related research fields such as domain generalization \cite{zhou2022domain}, transfer learning \cite{zhuang2020comprehensive} or few-shot learning \cite{wang2020generalizing}. 

Since a generalization across vehicle- and country-specific shifts is currently not feasible, the transferability, i.e. the transfer learning capabilities and possibilities of motion transformers, is of particular interest for practical application and scaling. Accordingly, a transfer learning study is being conducted based on the MTR framework, which is described in more detail in the following section.

\section{Transfer Learning Study}\label{transfer_learning}

While MTR and subsequent research show impressive results within a dedicated research setting, the ability to be adopted across different settings is crucial for its utility in various environments. Therefore, transferability is pivotal for making the technology widely accessible and ensuring a far-reaching impact. To investigate this capability, we designed a transfer learning study drawing from prior research in the field.

\subsection{General Setup}
The study aims at investigating the knowledge transfer from the Waymo Open Motion Dataset (WOMD), refered to as source dataset $\textit{D}_S$ to the constructed CarMaker Dataset (CMD), refered to as target dataset $\textit{D}_T$. However, before the dataset specifics and the comparison in the Table \ref{tab:dataset-comparison} are discussed, the task is considered more formally in accordance with \cite{zhuang2020comprehensive}. Hence, the datasets $\textit{D}$ represent observations of both a domain space and a task space, denoted as $\{(\mathcal{D}, \mathcal{T})\}$. Here, the domain space $\mathcal{D}=\{\mathcal{X}, P(X)\}$ encompasses an associated feature space $X$ and its corresponding marginal distribution $P(X)$, while the task space $\mathcal{T}=\{\mathcal{Y}, f\}$ is defined through the label space $\mathcal{Y}$ and the implicit function $f: \mathcal{X} \rightarrow \mathcal{Y}$ to be learned.

Accordingly, the transfer learning objective can be stated as improving the learning of the implicit target function $f_{T}$ by using source observations $\textit{D}_S=\{(\boldsymbol{x}_i,\boldsymbol{y}_i)\}^{n_S}_{i=1}$ of $\{(\mathcal{D}_S, \mathcal{T}_S)\}$ and target observations $\textit{D}_T=\{(\boldsymbol{x}_S,i,\boldsymbol{y}_S,i)\}^{n_T}_{i=1}$ of $\{(\mathcal{D}_T, \mathcal{T}_T)\}$, with $\boldsymbol{x}_{j,i} \in \mathcal{X}_j \subseteq \mathbb{R}^{d_{X_j}}$, $\boldsymbol{y}_{j,i} \in \mathcal{Y}_j \subseteq \mathbb{R}^{d_{y_j}}, j\in\{S,T\}$ and $n_s, n_T \in \mathbb{N}^{+}$ where $ 0 \ll n_T \ll n_S$. This highlights that unsupervised transfer learning ($n_T = 0$) is not part of our study, as generating a small, specific dataset is not only possible but also meaningful \cite{hanneke2019value} in the task at hand.

The datasets used as the foundation for the study are compared in Table \ref{tab:datasets}. The CMD is constructed using custom road map and agent state formatters to ensure that trajectories are represented in the same manner as in WOMD. Consequently, each trajectory consists of a sequence of states providing 3D center position, 2D velocity, heading and dimensions. Thus, the transfer learning setting at hand falls under the category of homogeneous inductive transfer learning.

\begin{table}[h]
    \centering
    \caption{Comparison of the Waymo Open Motion Dataset (WOMD) and the constructed CarMaker Dataset (CMD).}
    \label{tab:dataset-comparison}
    \begin{tabular}{p{3.5cm}p{2.cm}p{2.cm}}
    \toprule
    \textbf{Dataset} & \textbf{WOMD} & \textbf{CMD} \\
    \midrule
    Source / Target & Source & Target \\
    Environment Setting & US Roads & German Roads \\
    \midrule
    Number of Scenarios & 575,205 & 190,933 \\
    Duration of Each Scenario & 9 seconds & 9 seconds \\
    Trajectory Sampling Rate & 10Hz & 10Hz \\
    \midrule
    Training Split & $84.\overline6$\% (487,005) & 70\% (133,653) \\
    Validation Split & $7.\overline6$\% (44,100) & 15\% (28,640) \\
    Test Split & $7.\overline6$\% (44,100) & 15\% (28,640) \\
    \midrule
    Total Trajectories & 2,566,096 & 190,933 \\
    Agent Split (Veh./Cyc./Ped.) & 70/7/23\% & 100/0/0\% \\
    \bottomrule
    \end{tabular}
    \label{tab:datasets}
\end{table}

\subsection{Selected Methodologies}
Despite the existence of various differentiation criteria in transfer learning approaches, such as instance-based, reconstruction-based, discrepancy-based, feature-based, parameter-based, adversarial-based, and more \cite{pan2009survey, zhuang2020comprehensive}, only a limited number of methodologies have gained noteworthy relevance  \cite{wang2018deep, tan2018survey}. In the context of homogeneous inductive transfer learning and trajectory prediction, we assume closely related source and target tasks with shared commonalities in the domains. As a result, three methods: Multi-task learning (MTL), feature reuse (FR), and fine-tuning (FT) are selected for the transfer learning study.

\textit{1) Multi-task learning (MTL)}\footnote{As MTL aims to improve performance across tasks, it can also be used beyond pure transfer learning settings and is therefore sometimes differentiated in the literature \cite{zhuang2020comprehensive}. However, due to the inductive transfer and the application in the respective task, the classification as a transfer learning approach is appropriate.} is an inductive transfer learning approach \cite{pan2009survey} that leverages the relationships between related tasks, such as source and target tasks, during joint training to enhance performance across all tasks \cite{caruana1997multitask}. In essence, MTL mirrors a human learning intuition of benefiting from the knowledge gained in both source and target tasks \cite{zhang2018overview2}. Methodologically, it achieves this by incorporating shared knowledge via an inductive bias captured across both datasets. Furthermore, MTL has shown significant success in related fields such as computer vision \cite{zhang2014facial}, \cite{dai2016instance, Liu_2019_CVPR}, and natural language processing \cite{collobert2008unified, wu2015deep, worsham2020multi}. MTL also includes instance-based, feature-based and parameter-based approaches \cite{zhuang2020comprehensive}. With the latter, a further distinction can be made between hard and soft parameter sharing. A hard parameter sharing \cite{caruana1993multitask} based approach is selected in this study, as parameter-based approaches have proven to be very successful and hard sharing effectively counteracts overfitting \cite{baxter1997bayesian} thus preventing misleading conclusions.

\textit{2) Feature Reuse (FR)} in transfer learning involves a two-step training process. Initially, a model is trained on a source domain, followed by training on the target domain with preserved parameters and a domain-specific output stage \cite{wang2018deep}, \cite{raghu2019transfusion}. This methodology leverages insights from convolutional neural networks, where earlier layers capture general features, making them potentially relevant to the target problem \cite{oquab2014learning}. Feature reuse is particularly effective in domains like medical image classification, addressing limited annotated data and privacy concerns \cite{lopes2017pre, rajaraman2018pre}. In this study, we augment the model with additional self-attention blocks in both the encoder and decoder, serving as output modules after training on the source dataset. The original model parameters, trained on the source, remain frozen, while new parameters are randomly initialized and fine-tuned on the target dataset.

\textit{3) Fine-tuning (FT)} 
also involves a two-step training process. The first step consists of initial training on the source data, which serves as task-specific initialization. While the second step comprises subsequent training on the target data, which enables domain- and task-specific adaptations of the model parameters \cite{tan2018survey}. While FT involves adjusting the entire pre-trained model, selective fine-tuning of specific network components is conceivable \cite{shen2021partial}. In this context, the study also conducts fine-tuning individually on the 5,118,464 parameters of the encoder (FTE) and the 60,662,870 parameters of the decoder (FTD). Despite its apparent simplicity, fine-tuning has proven successful across a spectrum of related applications and architectures, especially up to large-scale models \cite{devlin2019bert, brown2020language}.

\subsection{Hyperparameters}

All training runs are performed for a certain number of epochs (30) using the PyTorch implementation of the AdamW optimizer \cite{loshchilov2019decoupled} with a fixed weight decay $\lambda=0.01$. The corresponding parameter update rule is given by
\begin{align}
    \label{eq:weightdecay}
    \boldsymbol{\theta}_{t} \leftarrow \boldsymbol{\theta}_{t-1} - \eta_{t} \Bigl(\frac{\hat{\boldsymbol{m}_{t}}}{\sqrt{\hat{\boldsymbol{v}_{t}}} + \epsilon} + \lambda \boldsymbol{\theta}_{t-1}\Bigr),
\end{align}
with schedulable learning rate $\eta_t$, the small scaler $\epsilon=10^{-8}$ to prevent division by zero, the bias corrected gradients $\hat{\boldsymbol{m}_t}$ and second moments $\hat{\boldsymbol{v}_t}$ as defined in \cite{kingma2014adam}. 

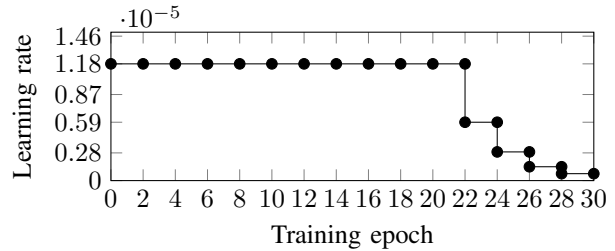
\begin{figure}[ht!]
\centering
\begin{tikzpicture}
    \begin{axis}[
        xlabel=Training epoch,
        ylabel=Learning rate,
        width=0.45\textwidth,
        height=0.2\textwidth,
        xmin=0, xmax=30,
        ymin=0, ymax=0.000015,
        xtick={0, 2, 4, 6, 8, 10, 12, 14, 16, 18, 20, 22, 24, 26, 28, 30},
        ytick={0.0, 0.0000028, 0.0000059, 0.0000087, 0.0000118, 0.0000146}
        ]
    \addplot[black, mark=*] plot coordinates {
                (0,  0.0000118) (2,  0.0000118) (4,  0.0000118) (6,  0.0000118) (8,  0.0000118)
                (10, 0.0000118) (12, 0.0000118) (14, 0.0000118) (16, 0.0000118) (18, 0.0000118)
                (20, 0.0000118) (22, 0.0000118) (22, 0.0000059) (24, 0.0000059) (24, 0.0000029)
                (26, 0.0000029) (26, 0.0000014) (28, 0.0000014) (28, 0.0000007) (30, 0.0000007)
    };
    \end{axis}
\end{tikzpicture}
\caption{Learning rate schedule across training epochs.}
\label{fig:learning_rate}
\end{figure}

While most of the hyperparameters adhere to the MTR training recommendations \cite{shi2022motion}, some adjustments are made. In particular, the batch size and the learning rate are selected differently due to the otherwise enormous memory requirements of the GPU. Accordingly, the batch size is reduced from 80 to one and the recommended initial learning rate $\eta_{\mathrm{rec. init.}}= 10^{-4}$ is adjusted to $ \eta_{\mathrm{init.}} = 1.18 \cdot 10^{-5}$ according to the rule of \cite{krizhevsky2014one}. Nevertheless, the schedule over the epochs corresponds to the MTR recommendation again. The resulting overall schedule for the learning rate across the training epochs is shown in Figure \ref{fig:learning_rate}.

\section{EVALUATION}\label{analysis}

This section first provides insights into the training and evaluation setup. Subsequently, both qualitative findings and quantitative results of the transfer learning study are presented. In addition, the frequently neglected computational training time is highlighted to provide information on possible trade-offs between computational time and performance.

\subsection{Training \& Evaluation Setup}
The training and evaluation setup is strongly correlated with the dataset generation process. To integrate the CarMaker simulation environment, five modules have been developed that are used for dataset generation, training and evaluation in various setups. These include a \textit{CarMaker extension}, a \textit{road map formatter}, an \textit{agent state formatter}, a \textit{model handler} and a graphical user interface refered to as \textit{GUI module}.

The \textit{Carmaker extension} is responsible for extracting and publishing simulation data such as time, position, heading and velocity via the User Datagram Protocol (UDP). The \textit{road map formatter} converts relevant road map data into the desired Waymo-like representation. The \textit{agent state formatter}, which focuses in particular on the agent state references and the sampling rate, serves a similar purpose for all observable agents. Accordingly, the simulation agent states are converted from absolute coordinates to ego-reference coordinates associated with the focal agent. For example, piecewise cubic Hermite interpolating polynomials (PCHIP) are used to resample and adapt the data to the standard input sampling rate of the motion transformer. 
 
While those modules are used for dataset generation, the \textit{model handler} provides an interface for loading, training and evaluating models as well as deploying the trained ones. Moreover, quantitative evaluation is conducted on the basis of the generated dataset, more precisely the respective test dataset split. In contrast, qualitative evaluation is enhanced using the \textit{GUI module}, that allows visual analysis. Furthermore, the GUI module enables the visual inspection of a trained model during deployment, e.g. by displaying motion predictions online. Similar to the generation of training data, CarMaker's stochastic traffic option can be used during operation to confront the ego vehicle.

All modules, the CarMaker simulation, the various trainings and evaluations of the respective methodologies selected in Section \ref{transfer_learning} run on a workstation PC configured as specified in Table \ref{tab:workstation-comparison}.
\begin{table}[h]
	\centering
	\caption{Configuration overview of the workstation used with regard to the central processing unit (CPU), the mainboard (MB), the  graphics processing unit (GPU), the random-access memory (RAM) and the solid-state drive (SSD).}
	\label{tab:workstation-comparison}
	\begin{tabular}{p{1.2cm}p{6.6cm}}
		\toprule
		\textbf{Component} & \textbf{Specification}   \\
		\midrule
		CPU & Intel Core i7-13700K, 8x 3.40 GHz, 8xE, 30MB L3 Cache \\
		MB & ASUS ROG STRIX Z790-F \\
		GPU & NVIDIA RTX A5000, 24GB GDDR6\\
		RAM & 64GB (2x 32GB Kit) DDR5-5200 CL40 \\
		SSD &2TB Samsung 980 Pro, M.2 PCIe 4.0 \\
		\bottomrule
	\end{tabular}
\end{table}

\subsection{Evaluation of Results}
The quantitative evaluation results are provided in Table \ref{tab:results}, using prediction scoring metrics as defined in Section \ref{motion_transformer_A}. Besides evaluation results of the methodologies outlined in Section \ref{transfer_learning}, baselines of source and target are provided, representing pure training on the corresponding dataset while being evaluated on both datasets. This way, performance is demonstrated with unmitigated shift. 

The results indicate that concerning mAP and minFDE on the source dataset, the source baseline performs the best. While MTL does not achieve improvement in mAP and minFDE on the source, its impact is evident in the metrics of minADE and miss rate. Overall, MTL did not bring about significant improvement on the source dataset, especially concerning mAP. Although these are interesting findings, performance on the target dataset is decisive for the evaluation of transfer learning, which is why this is examined in more detail below. 

The mAP of the target dataset shows that all transfer learning methods result in an improvement compared to the target baseline. However, even though MTL improves performance across all metrics compared to the target baseline, it is noticeable that MTL performs worse than the source baseline on the target. In comparison, the FR, FT, FTE and FTD provide improvement over both, the target and source baselines.

Particularly, fine-tuning performs the best across all metrics on the target. Moreover, FTE, FTD, and FR rank as the second-best, third-best, and fourth-best across all metrics, respectively. In the overall comparison between the target and source, it is evident that the performance is generally higher on the target dataset than on the source dataset. This is expected, as the simulation-based CarMaker target dataset has lower complexity than the advanced dataset provided by Waymo. 

\begin{figure}[ht]
\begin{subfigure}{.5\textwidth}
        \centering
        \includegraphics[width=0.6\linewidth]{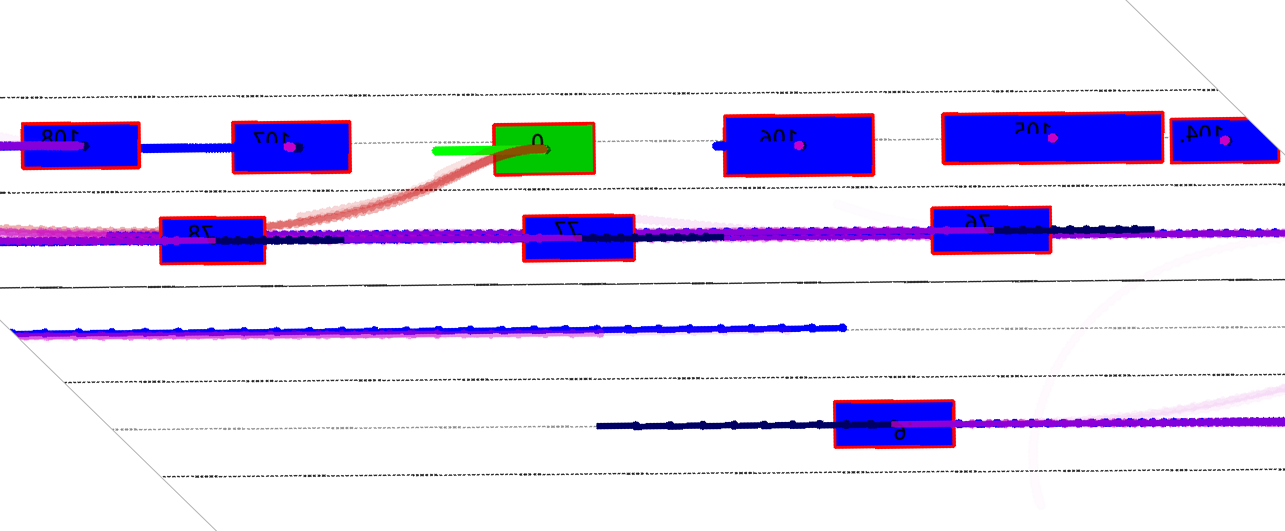}
        \caption{Source baseline model used on CMD.}
        \label{fig:T3S}
\end{subfigure}
\begin{subfigure}{.5\textwidth}
        \centering
        \includegraphics[width=0.6\linewidth]{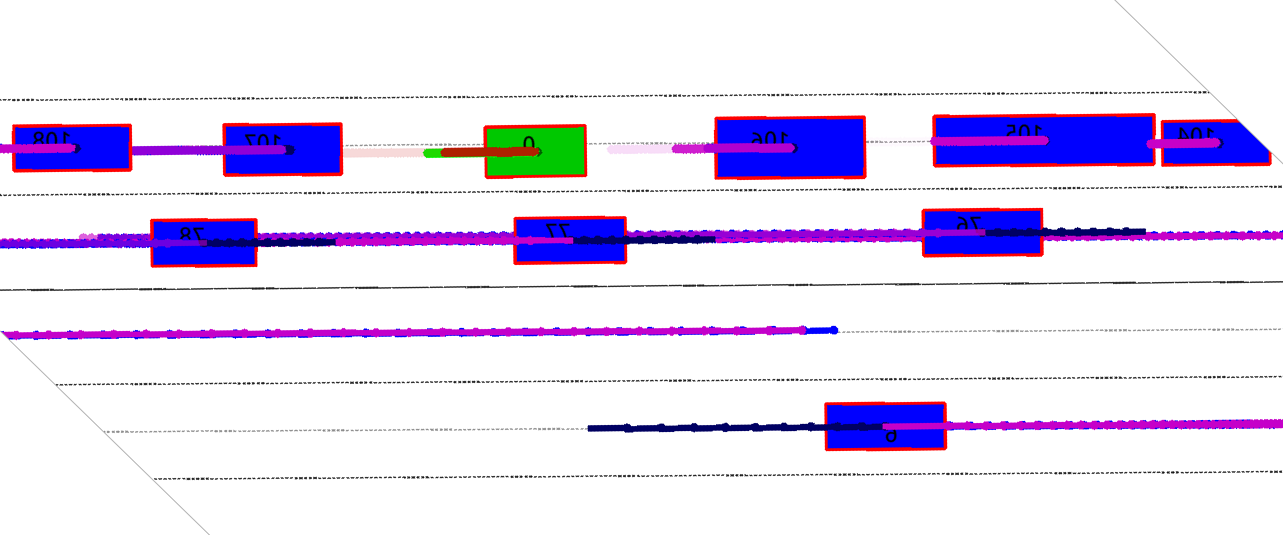}
        \caption{Fine-tuned model used on CMD.}
        \label{fig:T3F}
\end{subfigure}
\caption{Visual comparison between the source baseline model and the fine-tuned model on a CMD target dataset scenario. Here, a vehicle is proceeding forward. The focal vehicle and its ground truth trajectory are green, while its 6 predicted trajectories are red, with higher opacity indicated higher confidence. Neighboring vehicles and their ground truth trajectories are blue, while their predicted trajectories are purple.}
\label{fig:T3}
\end{figure}

\begin{figure}[ht]
\begin{subfigure}{.5\textwidth}
        \centering
        \includegraphics[width=0.59\linewidth]{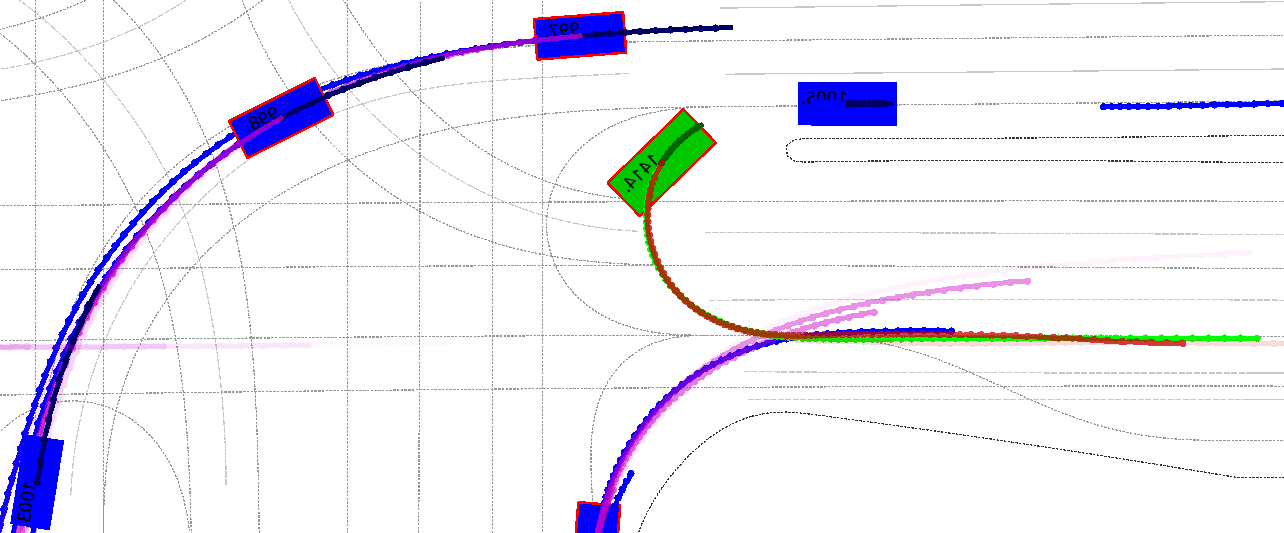}
        \caption{Source baseline model used on WOMD.}
        \label{fig:S2S}
\end{subfigure}
\begin{subfigure}{.5\textwidth}
        \centering
        \includegraphics[width=0.59\linewidth]{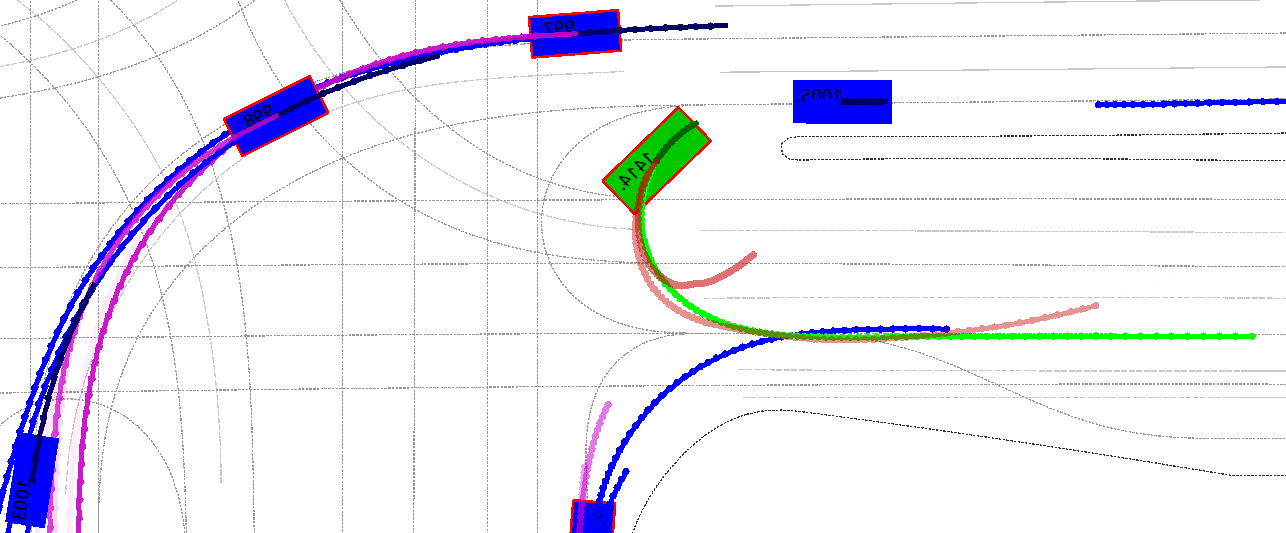}
        \caption{Fine-tuned model used on WOMD.}
        \label{fig:S2F}
\end{subfigure}
\caption{Visual demonstration of catastrophic forgetting of the fine-tuned model on an intersection scenario of the WOMD source dataset. Here, the vehicle begins to execute a U-turn. The focal vehicle and its ground truth trajectory are green, while its 6 predicted trajectories are red, with higher opacity indicates higher confidence. Neighboring vehicles and their ground truth trajectories are blue, while their predicted trajectories are purple.}
\label{fig:S2}
\end{figure}

In addition, the comparison in Figure \ref{fig:T3} provides a qualitative insight into the performance of FT. Moreover, it can be observed that FT model performs worse on the source dataset than the source baseline model. This is an indication of catastrophic forgetting \cite{mccloskey1989catastrophic}. A closer look at this phenomenon is illustrated in Figure \ref{fig:S2} using an example.

\begin{table*}[ht]
    \centering
    \caption{Evaluation of transfer learning study.}
    \begin{adjustbox}{max width=\textwidth}
    \begin{tabular}{l|cccc|cccc}
    % \toprule
    \multicolumn{1}{c}{} & \multicolumn{4}{|c|}{Source Dataset} & \multicolumn{4}{c}{Target Dataset} \\
    % \toprule \cmidrule(lr){1-1}
    \cmidrule(lr){2-5} \cmidrule(lr){6-9}
    & mAP $\uparrow$ & minADE $\downarrow$ & minFDE $\downarrow$ & MissRate $\downarrow$ & mAP $\uparrow$ & minADE $\downarrow$ & minFDE $\downarrow$ & missRate $\downarrow$ \\
    \midrule
    Target Baseline (TB) & 0.0917 & 4.6698 & 10.2989 & 0.6901 & 0.2078 & 2.4123 & 5.2653 & 0.4720 \\
    Source Baseline (SB) & \textbf{0.3968} & 0.8880 & \textbf{1.6266} & 0.1723 & 0.2947 & 1.5261 & 3.4756 & 0.3152 \\
    \midrule
    Multi-task learning (MTL) & 0.3290 & \textbf{0.8864} & 1.6550 & \textbf{0.1647} & 0.2391 & 1.9461 & 4.2792 & 0.3923 \\
    Fine-tuned (FT) & 0.1774 & 1.9257 & 3.4042 & 0.3973 & \textbf{0.6611} & \textbf{0.6508} & \textbf{1.2165} & \textbf{0.0782} \\
    Fine-tuned decoder (FTD) & 0.1887 & 1.6017 & 3.0359 & 0.3794 & 0.4413 & 0.9639 & 1.7298 & 0.1802 \\
    Fine-tuned encoder (FTE) & 0.1925 & 1.8823 & 3.4256 & 0.3914 & 0.4968 & 0.8308 & 1.6501 & 0.1669 \\
    Feature reuse (FR)& 0.1928 & 1.4524 & 2.7074 & 0.3588 & 0.3785 & 1.2275 & 2.4087 & 0.2684 \\
    \bottomrule
    \end{tabular}
    \end{adjustbox}
    \label{tab:results}
\end{table*}

\subsection{Computational Training Time}
Performance becomes irrelevant when a technique is too resource intensive to be practical. Therefore, all trainings conducted within this study are performed on a desktop computer equipped with a single GPU. More specifically a workstation PC according to Table \ref{tab:workstation-comparison} with a single NVIDIA A5000 GPU. All trainings are conducted by means of the training losses employed by the MTR Framework \cite{shi2022motion}, but training specifics such as datasets are defined according to the respective transfer learning methodologies to be analyzed.

\begin{figure}[ht!]
\centering
\begin{tikzpicture}
    \begin{axis}[
            ybar,
            xtick=data,
            width=0.45\textwidth,
            height=0.2\textwidth,
            %symbolic x coords={Source-only baseline, Target-only baseline, Multi-task learning, Fine-tuned, Fine-tuned decoder, Fine-tuned encoder, Feature Reuse},
            symbolic x coords={SB, TB, MTL, FT, FTD, FTE, FR},
            xticklabel style={rotate=45, anchor=east},
            ylabel={Train time [days]}
            %bar width=0.1\textwidth,
            %enlarge x limits=1,
                %xticklabels from table={\trainingtime}{ivl},
                %xticklabel style={text width=2cm, align=center},
        ]
        \addplot table [y=Duration]{\trainingtime};
    \end{axis}
\end{tikzpicture}
\caption{Total training duration for each model.}
\label{fig:training_time}
\end{figure}
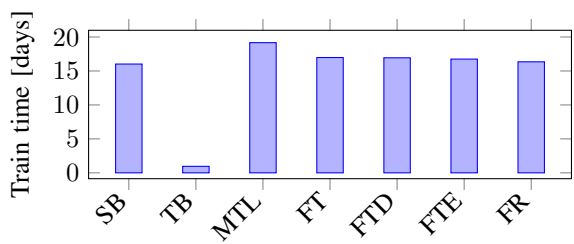

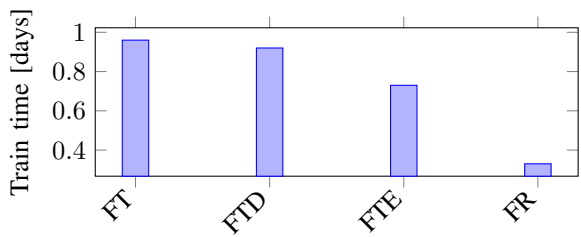
\begin{figure}[ht!]
\centering
\begin{tikzpicture}
    \begin{axis}[
            ybar,
            xtick=data,
            width=0.45\textwidth,
            height=0.2\textwidth,
            %symbolic x coords={Fine-tuned, Fine-tuned decoder, Fine-tuned encoder, Feature Reuse},
            symbolic x coords={FT, FTD, FTE, FR},
            xticklabel style={rotate=45, anchor=east},
            ylabel={Train time [days]}
            %bar width=0.1\textwidth,
            %enlarge x limits=1,
                %xticklabels from table={\trainingtime}{ivl},
                %xticklabel style={text width=2cm, align=center},
        ]
        \addplot table [y=Duration]{\tuningtime};
    \end{axis}
\end{tikzpicture}
\caption{Training duration on target data only.}
\label{fig:tuning_time}
\end{figure}

In the source baseline, target baseline, and MTL experiments, the model is trained on the entire training dataset all at once. In the FT, FTD, FTE and FR experiments, the models are first trained on the source dataset, which corresponds to the training of the source baseline, and thereafter the models are continued to be trained on the target dataset according to the respective method. In this way, these experiments can be built on a source baseline model and allow quick adaptation to different target datasets. This is illustrated in more detail in Figures \ref{fig:training_time} and \ref{fig:tuning_time}. While Figure \ref{fig:training_time} shows the total training time required for each model, Figure \ref{fig:tuning_time} illustrates the time required only for training on the target dataset of the respective methods. Since the target datasets are generally smaller, the corresponding training time is also smaller, which allows for better scalability across multiple target applications.

\pgfplotstableread[row sep=\\,col sep=&]{
    ivl & Duration \\
    %Source-only baseline & 16.02    \\
    %Target-only baseline & 0.961    \\
    %Multi-task learning  & 19.17    \\
    %Fine-tuned           & 16.98    \\
    %Fine-tuned decoder   & 16.94    \\
    %Fine-tuned encoder   & 16.75    \\
    %Feature Reuse        & 16.35    \\
    SB & 16.02    \\
    TB & 0.961    \\
    MTL & 19.17    \\
    FT  & 16.98    \\
    FTD & 16.94    \\
    FTE & 16.75    \\
    FR  & 16.35    \\
    }\trainingtime

\pgfplotstableread[row sep=\\,col sep=&]{
    ivl & Duration \\
    %Fine-tuned           & 0.96    \\
    %Fine-tuned decoder   & 0.92    \\
    %Fine-tuned encoder   & 0.73    \\
    %Feature Reuse        & 0.33    \\
    FT  & 0.96    \\
    FTD & 0.92    \\
    FTE & 0.73    \\
    FR  & 0.33    \\
    }\tuningtime

 \section{DISCUSSION}\label{discussion}

The results of the MTL provide evidence that the current state of the art in the area of vehicle motion prediction is not suitable for generalization across different domain and task spaces. In contrast, FT, FTD, FTE and FR provide better results in the target domain, but require the model to be re-adapted when the system is transferred back to the original source domain. 

With regard to the computational training time, it can be seen that FTE enables a significant reduction in training effort with only a slight reduction in performance and can therefore represent a practical approach that enables scalability. This is due to the significantly lower number of parameters for the encoder (8\%) compared to the number of parameters for the decoder (92\%). In addition to the significance of this study for transferability, the study also offers insights into the capabilities of the individual MTR components. Therefore, the following further research questions arise from this study. 

On the one hand, it can be assumed that the encoder is essential for the task of motion prediction and a targeted further development of the encoder could improve the generalization capability of the current state of the art. On the other hand, the results need to be examined on a larger scale, e.g. with datasets from different traffic environments or legal frameworks across countries and regions. This is particularly important to strengthen the findings and to promote the transfer of research results into real systems, but requires appropriate datasets.  The challenge of data generation is an important step and calls for cooperation between academia and industry in an open innovation culture.

\section{CONCLUSION}\label{conclusion}

The present transfer learning study demonstrates that among the examined methods, fine-tuning yields the most favorable results for motion transformer-based trajectory predictings. Multi-task learning, on the other hand, confirms that models for multiple settings currently do not appear to be realistic. Particularly noteworthy are the outcomes of encoder and decoder fine-tuning in relation to training duration. It is evident that fine-tuning the encoder results in a relatively minor performance loss, while noticeably reducing the training time. It suggests that encoder fine-tuning provides a scalable method for adapting to various settings, with relatively low performance losses. However, the applicability depends on the specific context, offering room for further research. Future research directions include the continuous development of the motion prediction methodology, for example with regard to the generalization capability of the encoder. Above all, however, the transfer capabilities can be investigated on a larger scale in different environments and legal frameworks in order to gain further insights and ultimately transfer the research on motion prediction to the real world.

\bibliographystyle{IEEEtran}
\bibliography{literature}

\end{document}